\documentclass{article}
\usepackage{spconf,amsmath,graphicx}
\usepackage{amsfonts}
\usepackage{cleveref}
\usepackage{url}
\usepackage{cite,balance}
\usepackage{textcomp}
\usepackage{amssymb}
\usepackage[ruled,vlined,linesnumbered]{algorithm2e}
\usepackage{multirow}
\usepackage{array}

\title{Improved Automated Machine Learning from Transfer Learning}
\name{Cat P. Le \qquad Mohammadreza Soltani \qquad Robert Ravier \qquad Vahid Tarokh \thanks{This work was supported by the Army Research Office grant No. W911NF-15-1-0479.}}
\address{Duke University}

\begin{document}
%
\maketitle
\begin{abstract}
In this paper, we propose a neural architecture search framework based on a similarity measure between some baseline tasks and a target task. We first define the notion of the task similarity based on the log-determinant of the Fisher Information matrix. Next, we compute the task similarity from each of the baseline tasks to the target task. By utilizing the relation between a target and a set of learned baseline tasks, the search space of architectures for the target task can be significantly reduced, making the discovery of the best candidates in the set of possible architectures tractable and efficient, in terms of GPU days. This method eliminates the requirement for training the networks from scratch for a given target task as well as introducing the bias in the initialization of the search space from the human domain. 
\end{abstract}

\section{Introduction}
\label{sec:introduction}

Most neural architecture search (NAS) methods focus on reducing the complexity of search by using a combination of explicit architecture search domains and specific properties of the given task at hand. These requirements make use of the existing architecture for a specific task less applicable to architecture search for other tasks. However, only a few of the above techniques~\cite{le2021task} have considered the similarity of tasks in the search space of architectures. This paper is motivated by a common assumption made in transfer and lifelong learning. In transfer learning literature~\cite{finn2016deep, niculescu2007inductive, luolabel, zamir2018taskonomy}, it is often believed that similar tasks can share similar architectures. In this light, using the architecture(s) of similar tasks as the potential architecture search space for a new task may reduce the dependence on prior knowledge and significantly accelerate the search of the final architecture. Building on this intuition, this paper proposes a NAS framework that learns an appropriate architecture for a given task based on its similarity to other tasks. A measure of similarity between pairs of tasks is proposed based on log-determinant of the Fisher information matrix of the loss function with respect to the parameters of the models under consideration. By definition, our measure of the task distance is asymmetric. Having an asymmetric measure is important since it might be easier to transfer the knowledge of a complex task to a simple task, but not the other way around. 

In our approach, we find the closest task to a target task in a given set of base tasks, and the architecture corresponding to the closest task is used to construct an architectural search space for the target task without requiring any prior knowledge from the human domain. Subsequently, the FUSE~\cite{le2021task} gradient-based search algorithm is applied to discover an appropriate architecture for the target task. Experimental results for $8$ classification tasks on MNIST~\cite{lecun2010mnist} and CIFAR-10~\cite{krizhevsky2009learning} datasets demonstrate the efficacy of the proposed approach over the other state-of-the-art methods. 

\section{Related Works}
\label{related}
Recent NAS techniques have been shown to offer competitive-to-superior performance to that of hand-crafted architectures. These techniques include approaches based on reinforcement learning (RL)~\cite{zoph2016neural}, and random search techniques~\cite{li2018massively,li2019random}. However, many of the existing NAS methods such as techniques based on the evolutionary algorithm~\cite{stanley2002evolving} often requires a lot of computational resources and thousands of GPU days to perform a search for a single task. Recently, differentiable search methods~\cite{liu2018darts,xie2018snas} have been proposed and, combined with random search methods and sampling sub-networks from one-shot super-networks~\cite{li2019random,cho2019one}, accelerated the underlying search time significantly. Additionally, random search~\cite{li2018massively,li2019random}, and network transformations~\cite{cai2018efficient} have shown to be very promising. Most recent techniques define the network architectures by stacking simple cells together~\cite{liu2018darts, xu2019pc}; as a result, the search space is also small and limited. Consequently, these techniques are often applicable to standard datasets, where the search space corresponding to each dataset is known. There are other approaches on combining task similarity with NAS~\cite{le2021neural,le2021task} in order to automatically generate a suitable search space for a given target task. Additionally, the task similarity is often been studied in transfer learning~\cite{finn2016deep, niculescu2007inductive, luolabel, zamir2018taskonomy}, and is used as regularization during learning to prevent catastrophic forgetting~\cite{kirkpatrick2017overcoming} in continual learning. However, the introduced measure of similarity in the above papers is assumed to be symmetric which limits their applicability in the neural architecture search.
\section{Neural Architecture Search Framework}
\label{sec:NAS}
In this section, we propose a neural architecture search (NAS) framework that considers the knowledge of the past learned tasks in order to construct a suitable search space for the incoming target task. This approach minimizes the requirement for prior knowledge from the human domain. Consider a set $B$ consisting of $K$ learned baseline tasks $T_i,$ and its corresponding dataset $X_i,$ denoted jointly by pairs $(T_i, X_i)$ for $i=1,2,\ldots,K$. Assume that for each baseline task, the best architecture and search space are known. In practice, the search for the best architecture for the baseline task can be perform using the random search method\cite{li2019random}. Below, a NAS framework, whose pseudo-code is provided in Algorithm~\ref{alg1} is presented for finding a well-performing architecture for a target task and data set pair $(T_t, X_t)$ based on the knowledge of architectures of these $K$ baseline learned tasks.
\begin{enumerate}
    \item \textbf{Task Similarity.} First, the dissimilarity of each learned task to the target task using the log-determinant task distance is computed. The closest baseline task based on the computed dissimilarities is returned.
    \item \textbf{Neural Architecture Search.} Next, a suitable search space for the target task is determined based on the closest task architecture. Subsequently, a search within this space is performed to find a well-performing architecture for the target task.
\end{enumerate}
\begin{algorithm}[t]
\SetKwInput{KwInput}{Input}         
\SetKwInput{KwOutput}{Output}       
\SetKwFunction{Distance}{Distance}
\SetKwFunction{FUSE}{FUSE}
\SetKwFunction{FMain}{Main}
\DontPrintSemicolon

\KwData{A set of baseline tasks: $B = \{(T_{1},X_{1}),...,(T_{K},X_{K})\}$}
\KwInput{$\varepsilon$-approx. network N, \# of candidates $C$, $\alpha = 1/|C|$, the incoming target $(T_t, X_t)$}
\KwOutput{Best architecture for the target task $t$}
    \SetKwProg{Fn}{Function}{:}{}
    \Fn{\Distance{$X_b, X_t, N_b, N_t$}}{
        Train $N_b$ with $X_b$\;
        Train $N_t$ with $X_t$\;
        Compute $F_{b,t}$ using $X_t$ on $N_b$ \;
        Compute $F_{t,t}$ using $X_t$ on $N_t$ \;
        \KwRet $\displaystyle d[b, t] =  \frac{1}{n} \bigg| \sum_{i} \log (\frac{\lambda_{b,t}^i+\sigma^2}{\lambda_{t,t}^i+\sigma^2}) \bigg|$\;
    }
    
    \SetKwProg{Fn}{Function}{:}{}
    \Fn{\FUSE{candidates $C$, data $X$}}{
        Define the relaxed output of C: $\displaystyle \Bar{c}(X) = \underset{c\in C}{\sum} \frac{\exp{(\alpha_c)}}{\underset{c'\in C}{\sum} \exp{(\alpha_{c'})}} c(X)$\;
        \While{$\alpha$ not converge}{
            Update $C$ by descending $\nabla_{w} \mathcal{L}_{tr}(w; \alpha, \Bar{c})$\;
            Update $\alpha$ by descending $\nabla_{\alpha} \mathcal{L}_{val}(\alpha; w, \Bar{c})$\;
        }
        \KwRet $c^* = \underset{c \in C}{\mathrm{argmin}}\ \alpha_c$\;
    }
    
    \SetKwProg{Fn}{Function}{:}{\KwRet}
    \Fn{\FMain}{
        \For{$b \in B$}{
            $d[b,t] = \Distance(X_b,X_t, N_b, N_t)$
        }
        Select closest task: $b^* = \underset{b \in B}{\mathrm{argmin}}\ d[b,t]$\;
        \While{criteria not met}{
            Sample $C$ candidates $\in$ search space $ S = S_{b^*} $\;
            $c^* = \FUSE \big(\{C, c^*\}, X_t \big)$\;
        }
        \KwRet best architecture $c^*$\;
    }
\caption{NAS with related search space}
\label{alg1}
\end{algorithm}

\subsection{Log-Determinant Task Distance}
As described above, we need to find the closest task to the target task between the learned baseline tasks. To this end, we define a dissimilarity measure between tasks based on the log-determinant of the Fisher information matrix. Let $\mathcal{P}_{N}(T,X)$ be a function that measures the performance of a given architecture $N$ on task $T$ with data $X$. An architecture $N$ is an $\varepsilon$-approximation network for $(T,X)$ if $\mathcal{P}_{N}(T,X) \geq 1 - \varepsilon$, for a given $0 < \varepsilon < 1$. These for example may be well-known hand-designed networks. Next, the empirical Fisher information matrices of the loss function of these $\varepsilon$-approximation networks are computed as follows:
\begin{equation}
    F = \frac{1}{M} \sum_{i=1}^M \nabla_\theta \log(p(x_i|\theta)) \nabla_\theta \log(p(x_i|\theta))^T,
\end{equation}
where $M$ is the number of data points, $\theta$ are the parameters of the neural network. Let $N_t$ be an $\varepsilon$-approximation network for the target task. Our goal is to evaluate how well the baseline approximation networks perform on the target task's data $X_t$. That is, the data from the target task is used to compute the empirical Fisher information matrices for all of the approximation networks. After computing these Fisher information matrices, we define the dissimilarity between tasks as follows. Let $b \in B$ and  $N_b$ be one of the baseline tasks and its corresponding $\varepsilon$-approximation network, respectively.  Let $F_{b, t}$ be the Fisher information matrix of $N_b$ with data $X_t$ from task $t$, and $F_{t, t}$ be the Fisher information matrix of $N_t$ with data $X_t$ from task $t$. We define the dissimilarity from task b to task t as:
\begin{multline}
    d[b, t] = \bigg| \frac{\log(\det(F_{b,t}+\sigma^2 * I_{n \times n}))}{n} \\
    - \frac{\log(\det(F_{t,t}+\sigma^2 * I_{m \times m}))}{m} \bigg|,
\end{multline}
where $I$ is the identity matrix, $\sigma$ is a pre-selected small constant, $n$ and $m$ are the number of parameters in $N_b$ and $N_t$, respectively. If $m=n$ or $N_b, N_t$ have a similar structure, then the distance can be expressed as:
\begin{align}{\label{eq-distance}}
    d[b, t] = \frac{1}{n} \bigg|  \sum_{i} \log (\frac{\lambda_{b,t}^i+\sigma^2}{\lambda_{t,t}^i+\sigma^2}) \bigg|,
\end{align}
where $\lambda^i$ is the i\textsuperscript{th} eigenvalue of the Fisher information matrix. This proposed dissimilarity is greater than or equal to $0$, with the distance $d=0$ indicating perfect similarity. Note that this dissimilarity is inherently asymmetric since it might be easier to transfer the knowledge of a comprehensive task to a simple task, but not vice versa.
\subsection{Neural Architecture Search}{\label{sec: fuse}}
Similar to recent NAS techniques~\cite{liu2018darts,le2021task}, our search space is defined by cells and skeletons. A cell is a densely connected graph of nodes, where nodes are connected by operations. The operations (e.g., identity, convolution) are set so that the dimension of the output is the same as that of the input. A skeleton is a structure consisting of multiple cells stacked together, forming a complete architecture. Next, the Fusion Search (FUSE) algorithm~\cite{le2021task} is applied to the reduced search space. The FUSE is a gradient-based search algorithm that evaluates all of the network candidates as a whole. It is based on the continuous relaxation of the outputs from all of the network candidates. Its goal is to search through all candidates without fully training them. 
Let $C$ be the set of network candidates from the given search space. For some $c \in C$ and data $X$, $c(X)$ denotes the output of the network candidate $c$. The relaxed output $\Bar{c}$ is the convex combination of the outputs from all candidates in $C$:
\begin{equation}{\label{softmax}}
    \Bar{c}(X) = \sum_{c\in C}\frac{\exp{(\alpha_c)}}{\sum_{c'\in C}\exp{(\alpha_{c'})}} c(X),
\end{equation}
where $\alpha_c$ is a continuous variable assigned to network $c$'s output. Next, we conduct the evaluation by jointly training the network candidates and optimizing their $\alpha$ coefficients. Let $X_{train}$, $X_{val}$ be the training and validation data. The training procedure is based on alternative minimization and can be divided into: (i) freeze $\alpha$ coefficients, jointly train network candidates and their weights, (ii) freeze network candidates' weights, update $\alpha$ coefficients. The best candidate in $C$ will be selected by: \(c^* = \arg\max_{c \in C} \alpha_c\).
To search through the entire search space, this process is repeated until certain criteria are met. Since the search space is restricted (reduced) only to the space of the most related tasks, the search algorithm is efficient and requires a reduced number of GPU-days.


\begin{figure}[t]
\begin{minipage}[b]{1\linewidth}
  \centering
  \centerline{\includegraphics[height=5.3cm]{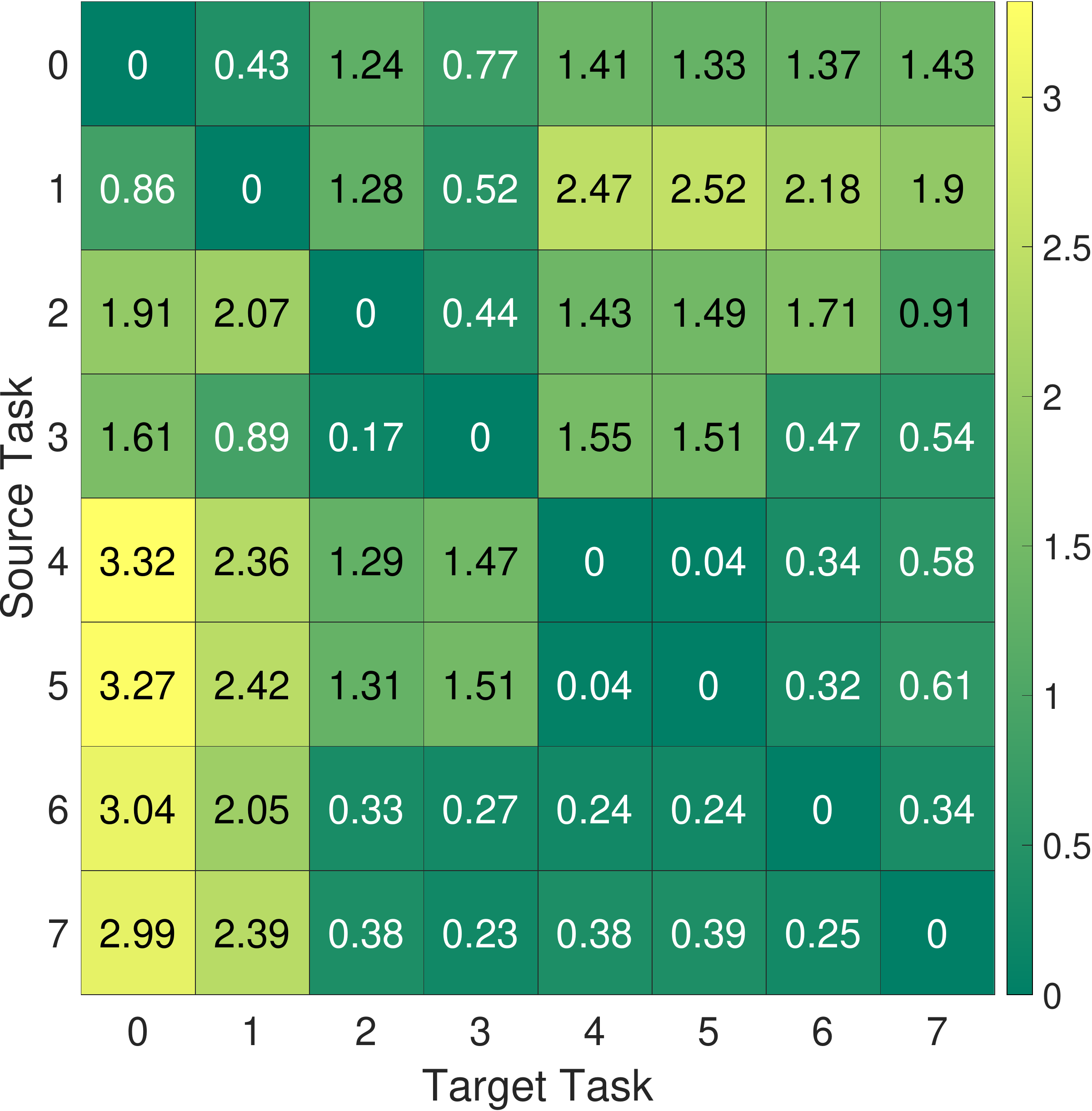}}
  \centerline{(a) Mean}\medskip
  \label{mean}
\end{minipage}
\begin{minipage}[b]{1\linewidth}
  \centering
  \centerline{\includegraphics[height=5.3cm]{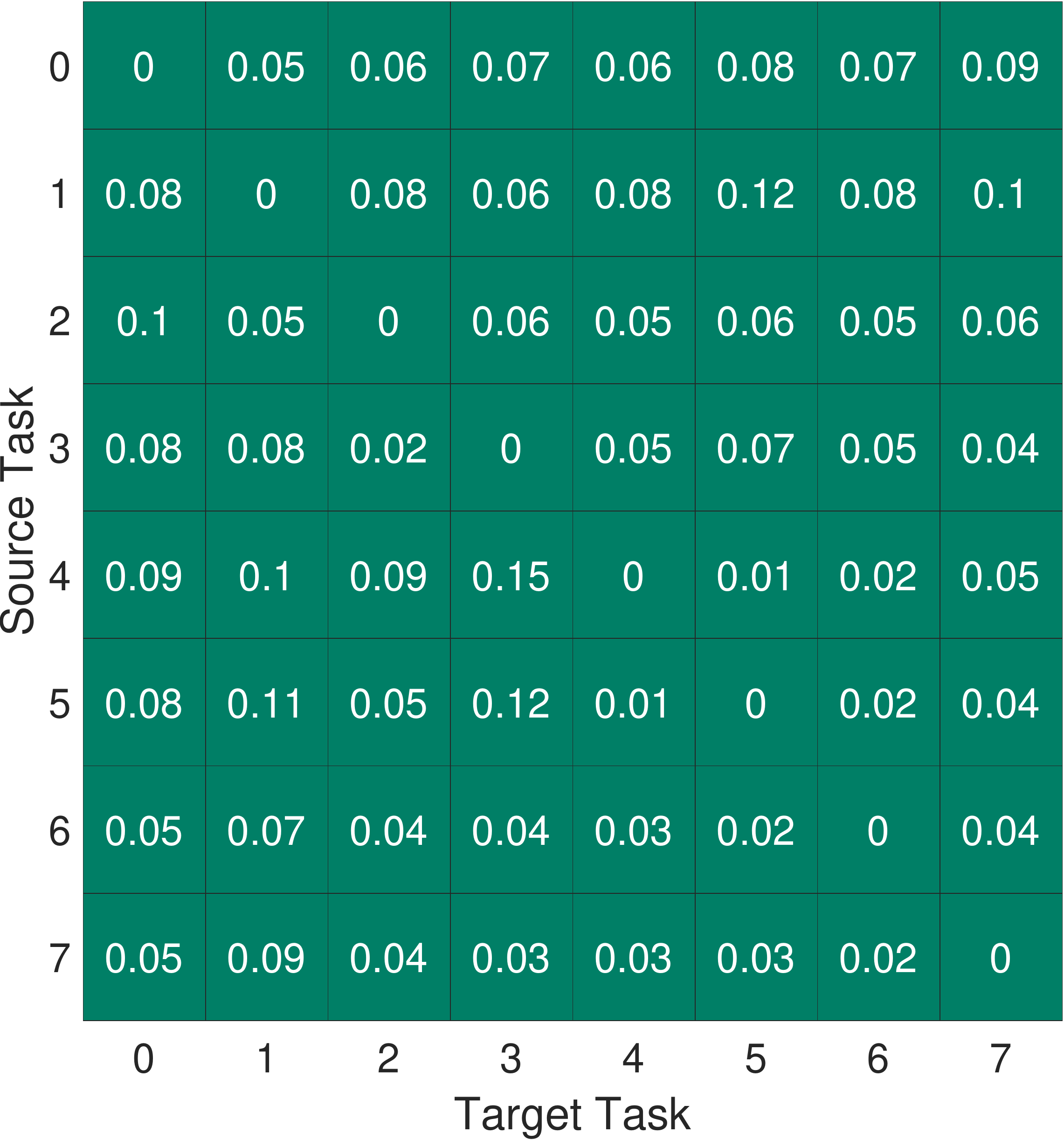}}
  \centerline{(b) Standard Deviation}\medskip
  \label{sig}
\end{minipage}

\caption{Distances between $8$ classification tasks from MNIST and CIFAR-10 datasets.}
\label{matrix}
\end{figure}

\section{Experimental Study}
\label{sec:experiment}

For our experiment, we initialize $4$ tasks in MNIST (i.e., Tasks $0-3$) and $4$ tasks in CIFAR-10 (i.e., Tasks $4-7$). 
Task $0$ and $1$ are defined as the binary classification tasks of detecting digits $0$ and $6$, respectively. Task $2$ is a binary classification of odd versus even digits. Task $3$ is the full $10$ digits classification. Task $4$ is a binary classification of three objects in CIFAR-10: automobile, cat, ship, i.e., the goal is to decide if the given input image consists of one of these three objects or not. Task $5$ is analogous to Task $4$ but with different objects (cat, ship, truck). Task $6$ is a multi-classification of four classes with labels bird, frog, horse, and anything else. Finally, Task $7$ is the standard $10$ object classification in CIFAR-10. We use the VGG-16 architecture~\cite{simonyan2014very} as the $\varepsilon$-approximation for both baseline and target tasks. To apply VGG-16 to both MNIST and CIFAR-10 tasks without architecture modification, we convert MNIST data from $1$ channel to $3$ channels and reshape images into $32 \times 32$ in order to match the dimension of CIFAR-10 data.

We first represent each task by an $\varepsilon$-approximation network by training the network with the training data corresponding to that task. Next, for each task, we compute $8$ empirical Fisher Information matrices using $8$ different datasets. For simplicity, the Fisher information matrices are approximated by only their diagonal entries. In other words, we only compute the diagonal entries of the Fisher Information Matrix, and set the non-diagonal entries to zero. In particular, for the network of size $n$ parameters, the number of Fisher information entries needed to be computed is $n$ instead of $n^2$ entries. This approach of simplification is often used in transfer learning~\cite{chen2018coupled}, and continual learning~\cite{kirkpatrick2017overcoming}. Next, we use Equation (\ref{eq-distance}) to compute the distance for each pair of tasks. Note that the distance from a task to itself will be zero since the difference between two Fisher Information matrices is zero. We repeat the experiment $10$ times with different initialization settings for the $\varepsilon$-approximation networks. The mean and standard deviation tables of the distance between tasks are illustrated in Figure~\ref{matrix}. The i\textsuperscript{th}  column of the mean table represents the average distance from other tasks to the target Task $i$. Our results suggest that two tasks from the same dataset (e.g., MNIST or CIFAR-10) are often more similar than tasks involving different datasets. It is perhaps surprising that the closest task to Task $3$ is Task $7$. Since other MNIST tasks are binary classification tasks, they do not intuitively appear as similar to the $10$-class classification, even though Task $3$ and Task $7$ are using different datasets.

We consider the problem of learning an architecture for Task $6$ from MNIST dataset, using the other aforementioned tasks as our baseline. From the sixth column of the mean table in Figure~\ref{mean}, it is observed that Task $7$ is the closest one to Task $6$. Thus, we apply cell structure and the operations of Task $7$ to generate a suitable search space for the target task. We use a cell structure, including $4$ nodes and $6$ operations. The list of related operations includes identity, sep-conv3x3, conv(7x1)(1x7), and maxpool3x3. The FUSE algorithm~\cite{le2021task} is then used to find the best architecture in this search space. Initially, five candidate architectures are randomly generated from this search space. At each iteration, the search algorithm evaluates these candidates and only saves the most promising architecture for the next iteration. The search stops when all criteria (e.g., a number of iterations or the best architecture converges) are met. The results in Table~\ref{table1} shows the best test accuracy of the optimal architecture found by our framework after $200$ iterations in comparison with state-of-art NAS methods (e.g., ENAS~\cite{pham2018efficient}, DARTS~\cite{liu2018darts}), a random search algorithm, and well-known handcrafted architectures, such as VGG~\cite{simonyan2014very}, ResNet~\cite{he2016deep}, DenseNet~\cite{huang2017densely}. The architecture produced by our framework is competitive with the hand-designed networks while having a significantly smaller number of parameters. When compared with the random search method, our approach achieves a higher-performing model with less running time. Our resulted architecture also outperforms ENAS~\cite{pham2018efficient} by a large margin in terms of classification performance, while having a much smaller number of parameters and requires fewer GPU days. When compare with DARTS~\cite{liu2018darts}, our architecture produces a competitive result, with a fewer number of parameters.
\begin{table}[t]
\caption{\label{table1}Comparison of the NAS performance with hand-designed classifiers and state-of-the-art  methods on Task $3$ in MNIST based on the discovered closest task, Task $7$.}
\begin{center}
\begin{tabular}{l|cc|cc|c}
\hline
\multicolumn{1}{l}{\bf Architecture}  &\multicolumn{1}{c}{\bf Accuracy} &\multicolumn{1}{c}{\bf Parameter} &\multicolumn{1}{c}{ \bf GPU}\\

\multicolumn{1}{l}{}  &\multicolumn{1}{c}{\bf (Task 3)} &\multicolumn{1}{c}{\bf (Millions)} &\multicolumn{1}{c}{ \bf days}\\
\hline
VGG-16~\cite{simonyan2014very}          & 99.55     & 14.72     & n/a\\ 
ResNet-18~\cite{he2016deep}             & 99.56     & 11.44     & n/a\\ 
DenseNet-121~\cite{huang2017densely}    & 99.61     & 6.95      & n/a\\
\hline
Random Search                           & 99.59     & 2.23      & 4\\ 
ENAS~\cite{pham2018efficient}           & 97.77     & 4.60      & 4\\
DARTS~\cite{liu2018darts}               & 99.51     & 2.37      & 2\\ 
\hline
LD-NAS (ours)                           & 99.67     & 2.28      & 2\\ 
\hline
\end{tabular}
\end{center}
\vspace{-.6cm}
\end{table}
Similarly, we consider Task $3$ from CIFAR-10 dataset as the incoming target task, and the other tasks as the baseline tasks. As observed from the mean table given in Figure~\ref{mean}, Task $7$ is the closest one to the target Task $3$. Table~\ref{table2} presents results indicating that the constructed architecture for Task $3$ has higher test accuracy with a significantly fewer number of parameters compared to hand-designed architectures. Our architecture also outperforms ENAS~\cite{pham2018efficient} by a large margin. When compare with DARTS~\cite{liu2018darts} and random search, our resulted network perform better in term of classification accuracy, and requires less search time. Our resulted model also has fewer parameters. Evidently, the proposed NAS framework can utilize the knowledge of the most similar task to efficiently find the optimal network architecture for the target task.

\begin{table}[t]
\caption{\label{table2}Comparison of the NAS performance with hand-designed classifiers and state-of-the-art  methods on Task $6$ in CIFAR-10 based on the discovered closest task, Task $7$.}
\begin{center}
\begin{tabular}{l|cc|cc|c}
\hline
\multicolumn{1}{l}{\bf Architecture} &\multicolumn{1}{c}{\bf Accuracy} &\multicolumn{1}{c}{\bf Parameter} &\multicolumn{1}{c}{ \bf GPU}\\

\multicolumn{1}{l}{} &\multicolumn{1}{c}{\bf (Task 6)} &\multicolumn{1}{c}{\bf (Millions)} &\multicolumn{1}{c}{ \bf days}\\
\hline
VGG-16~\cite{simonyan2014very}          & 86.75     & 14.72     & n/a\\ 
ResNet-18~\cite{he2016deep}             & 86.93     & 11.44     & n/a\\ 
DenseNet-121~\cite{huang2017densely}    & 88.12     & 6.95      & n/a\\
\hline
Random Search                           & 88.55     & 3.65      & 5\\ 
ENAS~\cite{pham2018efficient}           & 75.22     & 4.60      & 4\\
DARTS~\cite{liu2018darts}               & 90.11     & 3.12      & 2\\
\hline
LD-NAS (ours)                           & 90.87     & 3.02      & 2\\ 
\hline
\end{tabular}
\end{center}
\vspace{-.6cm}
\end{table}

\section{Conclusions}
\label{sec:conclusion}
A task similarity measure from a source task to a target task is given in this paper. By definition, this measure is asymmetric since applying knowledge of a comprehensive task into a simple task is easy but not vice versa. We apply this non-commutative distance measure in the neural architecture search framework. Using this task distance, a focus search space of architectures for a target task can be constructed using the closest task to the target from a set of learned baseline tasks. This reduces the complexity of the search space and increases the search's efficiency, results in the architecture with higher accuracy and a smaller number of parameters. 


\vfill
\pagebreak

\bibliographystyle{IEEEbib}
\bibliography{refs}

\end{document}